\documentclass[runningheads]{llncs}
\usepackage{graphicx}
\usepackage{xcolor}
\usepackage{multirow}
\usepackage{makecell} 
\usepackage{tabularx}
\usepackage{siunitx}
\usepackage{array}
\usepackage{colortbl}
\usepackage{xcolor}
\usepackage{cite} 
\usepackage{booktabs}
\usepackage{amsmath} 
\usepackage{paralist}
\definecolor{lightergray}{gray}{0.90}
\newcolumntype{M}[1]{>{\centering\arraybackslash}m{#1}} 
\usepackage[hidelinks]{hyperref}
\usepackage[T1]{fontenc}
\usepackage{cmtt}

\newcommand{\mypar}[1]{\smallskip\noindent\textbf{#1.}}

\newcommand{\myparthree}[1]{\smallskip\noindent\textbf{#1:}}
\newcommand{\myparfour}[1]{\smallskip\noindent\textit{#1}}

\newcommand\changed[1]{#1}

\newcommand{\prefix}{hd^k(\sigma)}
\newcommand{\prefixgraph}{G_{hd^k(\sigma)}}

\begin{document}
\title{PGTNet: A Process Graph Transformer Network for Remaining Time Prediction of Business Process Instances}
\titlerunning{PGTNet for Remaining Time Prediction}
\author{Keyvan Amiri Elyasi\inst{1} \orcidID{0009-0007-3016-2392} \and
Han van der Aa\inst{2} \orcidID{0000-0002-4200-4937}\and Heiner Stuckenschmidt\inst{1}\orcidID{0000-0002-0209-3859}}
\authorrunning{K. Amiri et al.}
\institute{Data and Web Science Group, University of Mannheim, Germany
\email{\{keyvan,heiner\}@informatik.uni-mannheim.de}
\and 
Faculty of Computer Science, University of Vienna, Austria
\email{han.van.der.aa@univie.ac.at}}
\maketitle              
\begin{abstract}
We present PGTNet, an approach that transforms event logs into graph datasets and leverages graph-oriented data for training Process Graph Transformer Networks to predict the remaining time of business process instances. PGTNet consistently outperforms state-of-the-art deep learning approaches across a diverse range of 20 publicly available real-world event logs. Notably, our approach is most promising for highly complex processes, where existing deep learning approaches encounter difficulties stemming from their limited ability to learn control-flow relationships among process activities and capture long-range dependencies. PGTNet addresses these challenges, while also being able to consider multiple process perspectives during the learning process.

\end{abstract}
\section{Introduction}
\label{sec:introduction}

Predictive process monitoring (PPM) aims to forecast the future behaviour of running business process instances, thereby enabling organizations to optimize their resource allocation and planning~\cite{tax_predictive_2017}, as well as take corrective actions~\cite{harl_explainable_2020}.
An important task in PPM is remaining time prediction, which strives to accurately predict the time until an active process instance will be completed. Precise estimations for remaining time are crucial for avoiding deadline violations, optimizing operational efficiency, and providing estimates to customers~\cite{tax_predictive_2017,navarin_lstm_2017}.

A variety of approaches have been developed to tackle remaining time prediction, with recent works primarily being based on deep learning architectures. In this regard, approaches using deep neural networks are among the most prominent ones~\cite{rama-maneiro_deep_2023}. However, the predictive accuracy of these networks leaves considerable room for improvement. In particular, they face challenges when it comes to capturing long-range dependencies~\cite{bukhsh_processtransformer_2021} and other control-flow relationships (such as loops and parallelism) between process activities~\cite{weinzierl_exploring_2022}, whereas they also struggle to harness information from additional process perspectives, such as case and event attributes~\cite{navarin_lstm_2017}.
Other architectures can overcome some of these individual challenges. For instance, the Transformer architecture can learn long-range dependencies \cite{bukhsh_processtransformer_2021},  graph neural networks (GNNs) can explicitly incorporate control-flow structures into the learning process \cite{weinzierl_exploring_2022}, and LSTM (long short-term memory) architectures can be used to incorporate (parts of) the data perspective~\cite{navarin_lstm_2017}.
However, so far, no deep learning approach can effectively deal with all of these challenges simultaneously.

Therefore, this paper introduces PGTNet, a novel approach for remaining time prediction that can tackle all these challenges at once. Specifically, our approach converts event data into a graph-oriented representation, which allows us to subsequently employ a neural network based on the general, powerful, scalable (GPS) Graph Transformer architecture \cite{rampasek_recipe_2022} to make predictions. Graph Transformers (GTs) have shown impressive performance in various graph regression tasks~\cite{kreuzer_rethinking_2021,dwivedi_graph_2022,rampasek_recipe_2022} and their theoretical expressive power closely aligns with our objectives: 
they can deal with multi-perspective data (covering various process perspectives) and can effectively capture long-range dependencies and recognize local control-flow structures. GTs achieve these latter benefits through a combination of local message-passing neural networks (MPNNs)\changed{\cite{pmlr-v70-gilmer17a}} and a global attention mechanism \cite{vaswani_attention_2017}. They employ sparse message-passing within their GNN blocks to learn local control-flow relationships among process activities, while their Transformer blocks attend to all events in the running process instance to capture the global context. 

We evaluated the effectiveness of PGTNet for remaining time prediction using 20  publicly
available real event logs. Our experiments show that our approach outperforms current state-of-the-art deep learning approaches in terms of accuracy and earliness of predictions. We also investigated the relationship between process complexity and the performance of PGTNet, which revealed that our approach particularly achieved superior predictive performance (compared to existing approaches) for highly complex, flexible processes.

The structure of the paper is outlined as follows: \autoref{sec:background} covers background and related work, \autoref{sec:preliminaries} presents preliminary concepts, \autoref{sec:graph transformers} introduces our proposed approach for remaining time prediction, \autoref{sec:evaluation} discusses the experimental setup and key findings, and finally, \autoref{sec:conclusion} summarizes our contributions and suggests future research directions.

\section{Background and Related work}
\label{sec:background}

This section briefly discusses related work on remaining time prediction and provides more details on Graph Transformers.

\mypar{Remaining time prediction}
Various approaches have been proposed for remaining time prediction, encompassing process-aware approaches relying on transition systems, stochastic Petri Nets, and queuing models, along with machine learning-based approaches~\cite{verenich_survey_2019}. In recent years, approaches based on deep learning have emerged as the foremost methods for predicting remaining time~\cite{rama-maneiro_deep_2023}. These approaches use different neural network architectures such as LSTMs~\cite{tax_predictive_2017,navarin_lstm_2017}, Transformers~\cite{bukhsh_processtransformer_2021}, and GNNs~\cite{duong_remaining_2023}.

Vector embedding and feature vectors, constituting the data inputs for Transformers and LSTMs, face a challenge in directly integrating control-flow relationships into the learning process. To overcome this constraint, event logs can be converted into graph data, which then acts as input for training a Graph Neural Network (GNN)~\cite{weinzierl_exploring_2022}. GNNs effectively incorporate control-flow structures by aligning the computation graph with input data. Nevertheless, they suffer from over-smoothing and over-squashing problems~\cite{kreuzer_rethinking_2021}, sharing similarities with LSTMs in struggling to learn long-range dependencies~\cite{rampasek_recipe_2022}. Moreover, existing graph-based predictive models face limitations due to the expressive capacity of their graph-oriented data inputs. Current graph representations of business process instances, either focus solely on the control-flow perspective \cite{harl_explainable_2020,venugopal_comparison_2021} or conceptualize events as nodes~\cite{weinzierl_exploring_2022,duong_remaining_2023}, leading to a linear graph structure that adversely impacts the performance of a downstream GNN.

\mypar{Graph Transformers}
Inspired by the successful application of self-attention mechanism in natural language processing, two distinct solutions have emerged to address the limitations of GNNs. The first approach unifies GNN and Transformer modules in a single architecture, while the second compresses the graph structure into positional (PE) and structural (SE) embeddings. These embeddings are then added to the input before feeding it into the Transformer network \cite{rampasek_recipe_2022}. Collectively known as Graph Transformers (GTs), both solutions aim to overcome the limitations of GNNs, by enabling information propagation across the graph through full connectivity \cite{kreuzer_rethinking_2021,rampasek_recipe_2022}. GTs also possess greater expressive power compared to conventional Transformers, as they can incorporate local context using sparse information obtained from the graph structure  \cite{dwivedi_graph_2022}.

Building upon this theoretical foundation, we propose to convert event logs into graph datasets to enable remaining time prediction using a Process Graph Transformer Network (PGTNet), as discussed in the remainder.

\section{Preliminaries}
\label{sec:preliminaries}
This section presents essential concepts that will be used in the remainder.

\mypar{Directed and attributed graphs}
A directed graph $G = (V,\mathcal{E})$ is defined by a set of nodes $V$, and a set of ordered pairs of nodes $\mathcal{E} \subseteq V \times V$ called directed edges. For an edge $\epsilon_{ij} = (v_{i},v_{j})\in \mathcal{E}$ pointing from $v_{i}$ to $v_{j}$, nodes $v_{i}\in V$ and $v_{j}\in V$ are called source and target nodes, respectively. Setting $n = |V|$, $G$'s adjacency matrix $A$ is a $n\times n$ matrix with $A_{ij} = 1$ if $\epsilon_{ij}\in \mathcal{E}$ and $A_{ij} = 0$ if $\epsilon_{ij}\notin \mathcal{E}$. An attributed graph is a graph that has node attributes in the form of a node feature matrix $X\in R^{n\times d}$, with $x_{v}\in R^d$ representing the feature vector of node $v$. It may also have edge attributes in the form of $\mathcal{Z}\in R^{m\times c}$, where $m = |\mathcal{E}|$ is the number of edges, and $z_{u,v}\in R^c$ denotes the feature vector of edge $(u,v)$.

\mypar{Events}
Let $\Gamma$ be the event universe (i.e., set of all possible event identifiers), $\mathcal{T}$ the time domain, $\mathcal{A}$ the finite set of process activities, $\mathcal{C}$ the set of all possible case identifiers, and $V_{1}$,$V_{2}$,...,$V_{m}$ sets of all possible values for data attributes $d_{1}$,$d_{2}$,...,$d_{m}$. An event $e$ is denoted by the tuple $e=(a,c,t,D)$. We assume that every event is characterized by mandatory properties, namely an activity identifier, a case identifier (i.e., the business process instance which the event belongs to), and a timestamp. Put it differently, there are $\Pi_{A}: \Gamma\rightarrow\mathcal{A}$, $\Pi_{C}:\Gamma\rightarrow\mathcal{C}$, and $\Pi_{T}: \Gamma\rightarrow\mathcal{T}$ functions that map an event $e$ to an activity, a case, and a timestamp: $\Pi_{A}(e)=a$, $\Pi_{C}(e)=c$, $\Pi_{T}(e)=t$. All other attribute-value pairs (e.g., transition life-cycle, organizational attributes, etc.) which may be associated to the event are denoted by $D \equiv\{(d_{1},v_{1}),(d_{2},v_{2}),...,(d_{m},v_{m})\}$ where $\forall v_{i}\in V_{i}$. Similar projection functions can be defined to extract values of specific attributes out of an event (e.g., $\Pi_{d_{i}}(e)=v_{i}$).  

\mypar{Traces, event log, event prefixes}
Let $\Gamma^\ast$ be the set of all possible sequences over $\Gamma$. A finite non-empty sequence of events $\sigma = <e_{1}, e_{2}, ..., e_{n}> \in\Gamma^\ast$, of length $|\sigma| = n$, is called a trace only and only if: 1) $\forall e_{i},e_{j}\in\sigma: \pi_{C}(e_{i}) = \pi_{C}(e_{j})$ (i.e., all events belong to the same case), 2) $\forall 1\leq i\leq j\leq n: \pi_{T}(e_{i})\leq \pi_{T}(e_{j})$ (i.e., the events are ordered by their timestamps). Each trace thus represents the execution of one business process instance. An event log refers to a collection of traces and is denoted as $L\subseteq \Gamma^\ast$, where every event appears only once within the entirety of the log. For the remaining time prediction problem, we also need to define partial traces, which we do through event prefixes. An event prefix of length $k$ consists of the first $k$ events of a trace and is denoted by $hd^k(\sigma)=<e_{1}, ..., e_{k}>$, where $k\in [1,n-1]$ is a positive integer number. 

\mypar{Problem statement}
Given an event log $L$ of completed traces and an event prefix $hd^k(\sigma)$ of a trace $\sigma = <e_{1}, e_{2}, ..., e_{n}>$, where $\sigma \notin L$ is an unseen trace of the same process as the traces of $L$, then the remaining time prediction problem is defined as using the traces from $L$ to learn a function $\theta_{L}$ that takes an unseen event prefix $hd^k(\sigma)$ and estimates the remaining time of the corresponding business process instance:

\begin{equation}
    \theta_{L}(hd^k(\sigma))\approx \pi_{T}(e_{n})-\pi_{T}(e_{k})\label{eq:RemTimeEstimate}
\end{equation}

\section{PGTNet for Remaining Time Prediction}
\label{sec:graph transformers}
To predict the remaining time of business process instances, we convert an event log into a graph dataset (see \autoref{subsec:transformation}), and use it to train a predictive model (see \autoref{subsec:GT architecture}). Once the model's parameters are learned, we can query the model to predict the remaining time of an active process instance based on its current partial trace. 

\subsection{Graph Representation of Event Prefixes}
\label{subsec:transformation}

To train a predictive model, we first turn an event log $L$ (consisting of traces of completed process instances) into a collection of event prefix-time tuples, with each tuple $(hd^k(\sigma), \pi_{T}(e_{n})-\pi_{T}(e_{k}))$ capturing an event prefix of length $k$ of a trace $\sigma \in L$, along with the time difference between $\sigma$'s final timestamp and the last timestamp of $\prefix$. Then, we establish a collection of directed attributed graphs $\mathcal{G}$, where each graph $G_{hd^k(\sigma)} \in \mathcal{G}$ encodes an event prefix $hd^k(\sigma)$, with its target attribute $\pi_{T}(e_{n})-\pi_{T}(e_{k})$.
This transformation is illustrated in \autoref{fig:Transformation_illustration}, displaying a snapshot of an event log for case ID `27583' and the corresponding graph representation of the event prefix of a length of 6. The details of this transformation procedure are as follows:

\begin{figure}[ht]
    \centering    \includegraphics[width=1\linewidth]{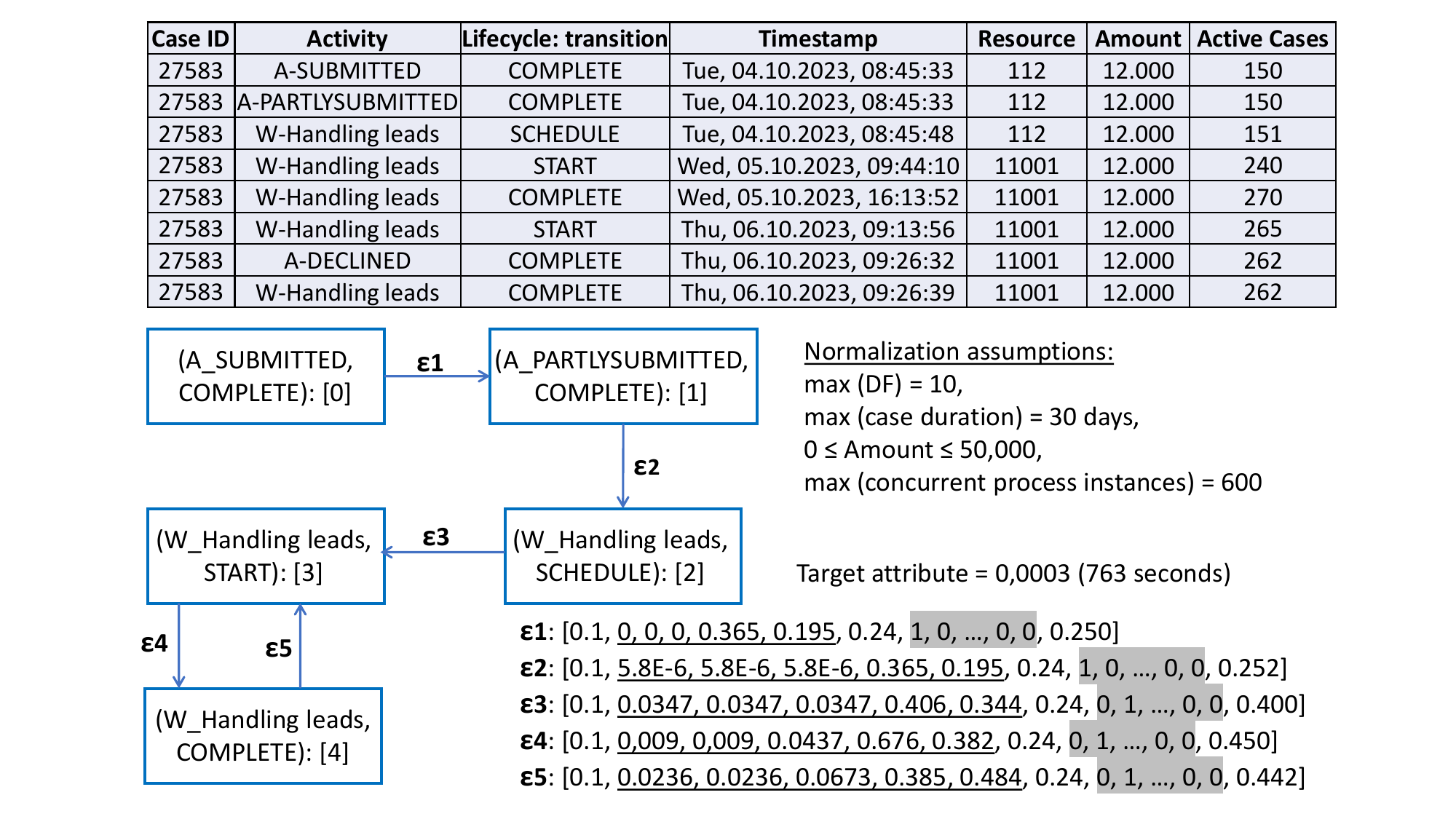}
    \caption{Graph representation of an event prefix of length of 6, Case ID= `27583'.}
    \label{fig:Transformation_illustration}
\end{figure}

\mypar{Nodes}
We create a node in $\prefixgraph$ for each \changed{event class, i.e., each} unique combination of an activity and life-cycle attribute (if any) contained in prefix $\prefix$. Each node has a numeric identifier, such as (`A-SUBMITTED', `COMPLETE'): [0] in \autoref{fig:Transformation_illustration}. 
Using event classes as graph nodes results in graphs with relatively few nodes, even for lengthy prefixes.

\mypar{Edges and edge weights} We create an edge in $\prefixgraph$ for each directly-follows (DF) relation in $\prefix$, while the edge's weight indicates the number of its occurrences within $\prefix$. These edge weights are normalized by the maximum number of occurrences of any directly-follows relation within the available training prefixes. 
In \autoref{fig:Transformation_illustration}, this normalization coefficient (\texttt{max(DF)}) is equal to 10, resulting in edge weights of 0.1 (the first feature in the edge's vector).

\mypar{Edge features}
To enhance the expressive capacity of the graph representation, we incorporate additional features into the edge feature vector:
\begin{compactitem}
  \item We use five different temporal features per edge. These  include the total duration ($t_{1}$) and duration of the last occurrence ($t_{2}$) of the DF relation represented by the edge. Similar to other works~\cite{tax_predictive_2017}, we also incorporate distances between the timestamp of the target node and the start of the case ($t_{3}$), start of the day ($t_{4}$), and start of the week ($t_{5}$) for the latest occurrence of the DF relation. While $t_{1}$, $t_{2}$, and $t_{3}$ are normalized by the largest case duration in the training data, $t_{4}$ and $t_{5}$ are normalized by the duration of days and weeks, respectively. In \autoref{fig:Transformation_illustration}, the temporal features are underlined in the feature vectors  of the edges.
  \item We encode case attributes and event attributes of the target node (for the last occurrence of the DF relation in $\prefix$). For numerical and categorical attributes, we use min-max normalization and one-hot encoding, respectively. In \autoref{fig:Transformation_illustration}, one-hot encoding of the categorical attribute `Resource' is highlighted with a grey shadow, while the feature before this gray part (always 0.24 in the example) captures the case attribute `Amount'. 
  \item To account for the overall workload of the process at a given time, we capture the number of active cases at the timestamp of the target node (for the last occurrence of the DF relation). This feature is normalized by the maximum number of concurrent process instances observed in the training data.  
\end{compactitem}

\noindent \changed{Note that we encode this information as edge features, rather than on the nodes, in order to preserve the simplicity of the node semantics. In this way, PGTNet can also deal with event logs with a large number of event classes, achieved by employing an embedding layer.}

\mypar{Target attribute}
For each graph $G_{hd^k(\sigma)} \in \mathcal{G}$, its target attribute $\pi_{T}(e_{n})-\pi_{T}(e_{k})$ is normalized by the longest case duration in the training data. 

\subsection{Training PGTNet to Predict Remaining Time}
\label{subsec:GT architecture}
Once an event log is converted into a graph dataset, it can be used to train PGTNet to learn function $\theta_L$ in \autoref{eq:RemTimeEstimate} in an end-to-end manner. We specifically approach the remaining time prediction problem as a graph regression task, using L1-Loss (mean absolute error between predictions and ground truth remaining times). Model training employs the backpropagation algorithm to iteratively minimize the loss function. For
this, we adopt the GPS Graph Transformer recipe \cite{rampasek_recipe_2022} as the underlying architecture of PGTNet.

\begin{figure}[ht]
    \centering    \includegraphics[width=1\linewidth]{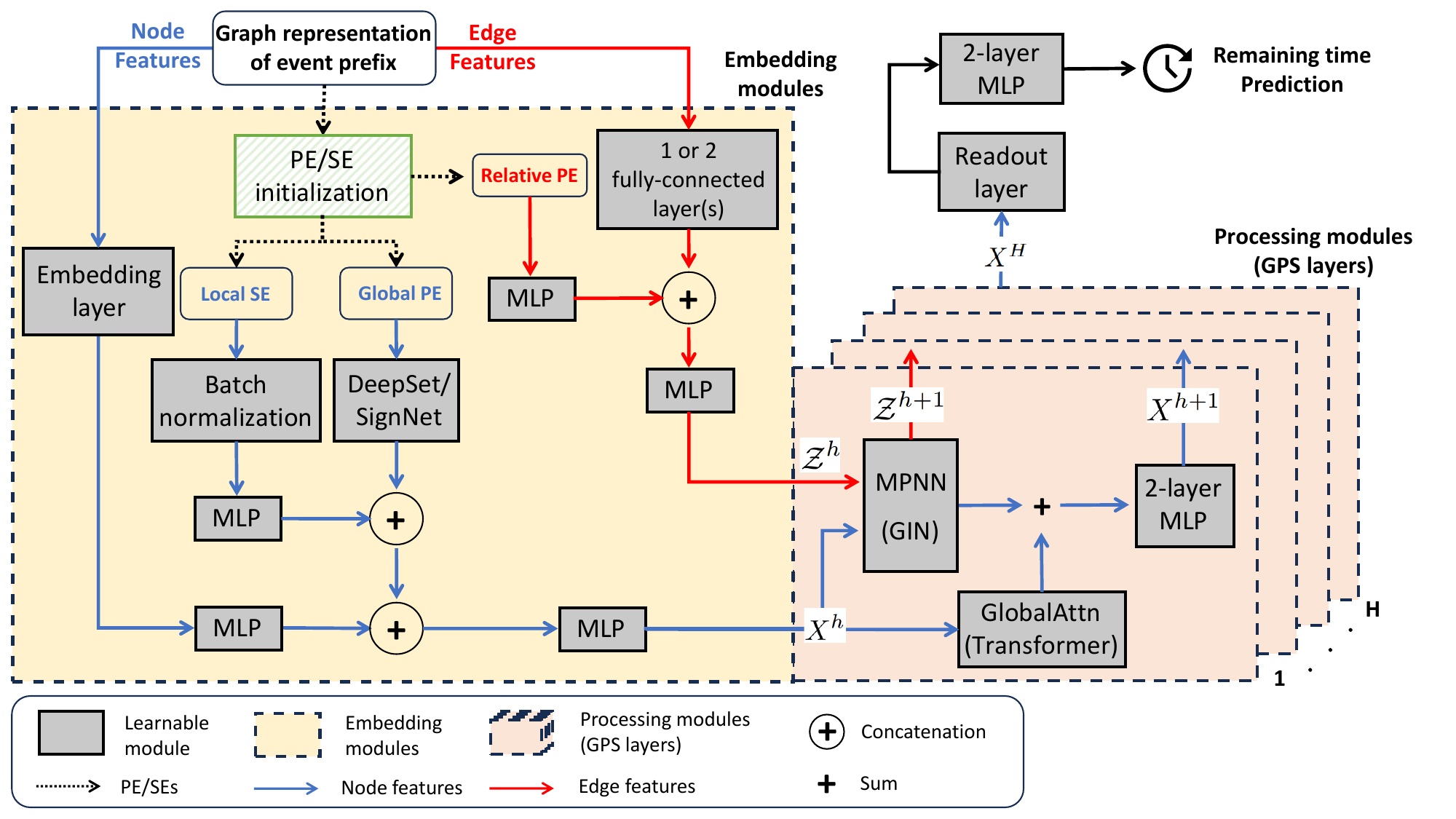}
    \caption{PGTNet architecture: based on the GPS Graph Transformer recipe \cite{rampasek_recipe_2022}. Paths to process node and edge features are specified by blue and red colors, respectively.}
    \label{fig:GPS_model}
\end{figure}

\mypar{PGTNet architecture}
PGTNet's architecture comprises embedding and processing modules, as shown in \autoref{fig:GPS_model}.

\myparfour{Embedding modules} have two main functionalities:
\begin{compactitem}
    \item They map node and edge features into continuous spaces. To ensure that similar event classes are closer in the embedding space, an embedding layer is used to map integer node features into a continuous space. We use fully-connected layer(s) to compress edge features into the same hidden dimension and address the challenges arising from high-dimensional data attributes. 
    \item They compress the graph structure into multiple positional and structural encodings (PE/SE), and seamlessly incorporate these PE/SEs into node and edge features \cite{rampasek_recipe_2022}. This integration is achieved through diverse PE/SE initialization strategies and the utilization of several learnable modules, including MLPs (multi-layer perceptron) and batch normalization layers, as illustrated in \autoref{fig:GPS_model}.
\end{compactitem}

\changed{Assuming a graph $G = (V,E)$ with adjacency matrix $A$, node feature matrix $\tilde{X}^{0}$, and edge feature matrix $\tilde{\mathcal{Z}}^{0}$, embedding modules are described by \autoref{eq:PE-SE}.} 

\begin{equation}
    X^{0}, \mathcal{Z}^{0} = \mathbf{f}_{PE/SE} (\tilde{X}^{0}, \tilde{\mathcal{Z}}^{0}, A) \label{eq:PE-SE}
\end{equation}

\changed{Graph Transformers vary in their choice of function $\mathbf{f}_{PE/SE}$, which is often a neural network with learnable parameters.} Further insights into the PE/SEs is provided in the subsequent discussion of the design space for PGTNet.

\myparfour{Processing modules} consist of hybrid layers combining MPNN and Transformer blocks. Each of the $H$ layers, also referred to as GPS layers, computes a hidden representation of nodes and edges ($X^{h+1}$, $\mathcal{Z}^{h+1}$) based on the node and edge embedding from the previous layer ($X^{h}$, $\mathcal{Z}^{h}$), and the adjacency matrix $A$, as summarized in Equations \ref{eq:GPS-layer}, \ref{eq:MPNN-block}, \ref{eq:Transformer-block}, and \ref{eq:GPS-MLP} below. Parallel computations in MPNN and Transformer blocks aim to strike a balance between local message passing and global attention mechanisms and resolve over-smoothing and over-squashing problems \cite{rampasek_recipe_2022}.\\

\begin{subequations}
\label{eq:GPS}
\begin{align}
    X^{h+1}, \mathcal{Z}^{h+1} &= GPS^{h}(X^{h}, \mathcal{Z}^{h},A) \quad \forall h \in \{0,1,...,H-1\}\ \label{eq:GPS-layer}\\
    X_{M}^{h+1}, \mathcal{Z}^{h+1} &= MPNN^{h}(X^{h}, \mathcal{Z}^{h},A) \quad \forall h \in \{0,1,...,H-1\}\ \label{eq:MPNN-block}\\
    X_{T}^{h+1} &= GlobalAttn^{h}(X^{h}) \quad \forall h \in \{0,1,...,H-1\}\ \label{eq:Transformer-block}\\
    X^{h+1} &= MLP^{h}(X_{M}^{h+1}+X_{T}^{h+1}) \quad \forall h \in \{0,1,...,H-1\}\ \label{eq:GPS-MLP}
\end{align}
\end{subequations}

After obtaining node embeddings $X^{H}$ from the last GPS layer, they are aggregated in the readout layer to derive a graph-level representation that is subsequently fed into a 2-layer MLP to predict the remaining time \changed{(see \autoref{eq:readout-head})}. \changed{Since a  simple permutation invariant function (e.g., mean or sum) is used for the readout function in \autoref{eq:readout-head}, the readout layer accommodates varying node counts. Therefore,} PGTNet avoids the need for zero-padding \cite{navarin_lstm_2017,bukhsh_processtransformer_2021} and its computational overhead. 

\begin{equation}
    y_{G} = MLP(readout(X^{H})) \label{eq:readout-head}
\end{equation} 

Note that edge features are solely processed by MPNN blocks and are not utilized by Transformer blocks or in obtaining the graph-level representation.

\mypar{Design space for PGTNet}
The modular design of the GPS Graph Transformer recipe  offers flexibility in choosing various types of positional/structural encodings (PE/SEs) and MPNN/Transformer blocks.

PE/SEs aim to enhance positional encoding for Transformer blocks \cite{kreuzer_rethinking_2021}, and enable GNN blocks to be more expressive \cite{dwivedi_graph_2022}. The compression of graph structure into PE/SEs can be achieved through the utilization of various initialization strategies (PE/SE initialization in \autoref{fig:GPS_model}). Notably, Laplacian eigenvector encodings (LapPE) \cite{kreuzer_rethinking_2021} furnishes node embedding with information about the overall position of the event class within the event prefix (global PE), while it enhances edge embedding with information on distance and directional relationships between nodes (relative PE). Random-walk structural encoding (RWSE) \cite{dwivedi_graph_2022} incorporates local SE into node features, facilitating the recognition of cyclic control-flow patterns among event classes. Graphormer employs a combination of centrality encoding (local SE) and edge encoding (relative PE) to enhance both node and edge features  \cite{ying_transformers_2021}.

Additionally, a range of learnable modules for processing PE/SEs can be integrated into PGTNet as highlighted in \cite{rampasek_recipe_2022}. These design options include simple MLPs as well as more advanced networks such as DeepSet \cite{zaheer_deep_2017} and SignNet \cite{lim_sign_2022}. Lastly, while it is possible to use various MPNN and global attention blocks within each GPS layer \cite{rampasek_recipe_2022}, we exclusively  used the graph isomorphism network (GIN) \cite{hu_strategies_2020} and conventional transformer architecture \cite{vaswani_attention_2017}. Further details regarding our policy for designing PGTNet are elaborated upon in \autoref{subsec:experiment}.

\section{Evaluation}
\label{sec:evaluation}

This section presents the experiments used to evaluate the performance of PGTNet for remaining time prediction.
\autoref{tab1} summarizes the characteristics of the 20 publicly available event logs used as a basis for this.
In the remainder, \autoref{subsec:experiment} describes the experimental setup, followed by the results in \autoref{subsec:results}.
Our employed implementation and additional results are available in our project's public repository.\footnote{\url{https://github.com/keyvan-amiri/PGTNet}}

\begin{table}[ht]
\caption{Characteristics of the employed event logs (time-related attributes in days).}\label{tab1}
\centering
\begin{tabular}{
    l
    S[table-format=6.0, table-column-width=1.2cm]
    S[table-format=6.0, table-column-width=1.4cm]
    S[table-format=3.0, table-column-width=1.1cm]
    S[table-format=4.0, table-column-width=1.3cm]
    S[table-format=2.2, table-column-width=0.9cm]
    S[table-format=3.0, table-column-width=0.9cm]
    S[table-format=3.1, table-column-width=1.1cm]
    S[table-format=4.1, table-column-width=1.3cm]}
\hline
\multicolumn{1}{m{1.8cm}}{\multirow{2}{*}{Event log}} & \multicolumn{1}{c}{\multirow{2}{*}{Cases}} & \multicolumn{1}{c}{\multirow{2}{*}{Events}} & \multicolumn{1}{c}{\multirow{2}{*}{\parbox{1.1cm}{\centering Event Classes}}} & \multicolumn{1}{m{1.2cm}}{\multirow{2}{*}{Variants}} & \multicolumn{2}{c}{Case length} &  \multicolumn{2}{c}{Case duration}\\
\cline{6-7} \cline{8-9}
 &  &  &  &  & \multicolumn{1}{c}{Avg.} & \multicolumn{1}{c}{Max} &  \multicolumn{1}{c}{Avg.} & \multicolumn{1}{c}{Max}\\
\hline
Env.permit & 1434 & 8577 & 27 & 116 & 5.98 & 25 & 5.4 & 275.8\\
Helpdesk & 4580 & 21348 & 14 & 226 & 4.66 & 15 & 40.9 & 60.0\\
BPIC12 & 13087 & 262200 & 36 & 4366 & 20.04 & 175 & 8.6 & 137.2\\
BPIC12W & 9658 & 170107 & 19 & 2643 & 17.61 & 156 & 11.7 & 137.2\\
BPIC12C & 13087 & 164506 & 23 & 4336 & 12.57 & 96 & 8.6 & 91.5\\
BPIC12CW & 9658 & 72413 & 6 & 2263 & 7.50 & 74 & 11.4 & 91.0\\
BPIC12O & 5015 & 31244 & 7 & 168 & 6.23 & 30 & 17.2 & 89.6\\
BPIC12A & 13087 & 60849 & 10 & 17 & 4.65 & 8 & 8.1 & 91.5\\
BPIC20I & 6449 & 72151 & 34 & 753 & 11.19 & 27 & 86.5 & 742.0\\
BPIC20D & 10500 & 56437 & 17 & 99 & 5.37 & 24 & 11.5 & 469.2\\
Sepsis & 1050 & 15214 & 16 & 846 & 14.49 & 185 & 28.5 & 422.3\\
Hospital & 100000 & 451359 & 18 & 1020 & 4.51 & 217 & 127.2 & 1035.4\\
BPIC15-1 & 1199 & 52217 & 398 & 1170 & 43.55 & 101 & 95.9 & 1486.0\\
BPIC15-2 & 832 & 44354 & 410 & 828 & 53.31 & 131 & 160.3 & 1326.0\\
BPIC15-3 & 1409 & 59681 & 383 & 1349 & 42.36 & 123 & 62.2 & 1512.0\\
BPIC15-4 & 1053 & 47293 & 356 & 1049 & 44.91 & 115 & 116.9 & 927.0\\
BPIC15-5 & 1156 & 59083 & 389 & 1153 & 51.11 & 153 & 98.0 & 1344.0\\
BPIC13I & 7554 & 65533 & 13 & 2278 & 8.68 & 123 & 12.1 & 771.4\\
BPIC13C & 1487 & 6660 & 7 & 327 & 4.48 & 35 & 178.9 & 2254.0\\
Traffic fines & 150370 & 561470 & 11 & 231 & 3.73 & 20 & 341.6 & 4372.0\\
\hline
\end{tabular}
\end{table}

\subsection{Experimental Setup}
\label{subsec:experiment}

\mypar{Data preprocessing}
We filter out traces with fewer than 3 events, since our approach requires an event prefix of at least length 2 to make a prediction. Aside from that, we do not apply any preprocessing to the available event logs.

\mypar{Benchmark approaches}
We compare our approach against four others:
\begin{compactitem}
    \item \textit{DUMMY}: A simple baseline that predicts the average remaining time of all training prefixes with the same length $k$ as a given prefix.

    \item \textit{DALSTM}~\cite{navarin_lstm_2017}: An LSTM-based approach that was recently shown to have superior results among LSTMs used for remaining time prediction~\cite{rama-maneiro_deep_2023}.

    \item \textit{ProcessTransformer}~\cite{bukhsh_processtransformer_2021}: A transformer-based approach designed to overcome LSTM's limitations \changed{in capturing long-range dependencies}.

   \item \changed{\textit{GGNN}~\cite{duong_remaining_2023}: An approach that utilizes gated graph neural networks to incorporate control-flow relationships into the learning process. It employs gated recurrent unit (GRU) within its MPNN layers, enabling the learning of the sequential nature of graph nodes.}
   
\end{compactitem}

\mypar{Data split}
There is no consensus on the data split for predictive process monitoring. Some papers employed a chronological holdout split \cite{tax_predictive_2017, bukhsh_processtransformer_2021}, while others opted for cross-validation \cite{evermann_predicting_2017, polato_data-aware_2014, rama-maneiro_deep_2023}. The holdout split maintains chronological order but may introduce instability due to end-of-dataset bias. To avoid such instability, additional data preprocessing, as suggested by \cite{weytjens_creating_2022}, is necessary.

As our benchmark approaches were not trained with such preprocessing, we avoided additional steps and chose a 5-fold cross-validation data split (CV=5) to enhance model's robustness against the end-of-dataset bias.  We randomly partition the dataset into 5 folds  where each fold serves as a test set once, and the remaining four folds are used as training and validation sets. For completeness, we also report results obtained using holdout data splits in our supplementary GitHub repository.

\mypar{Prefix establishment}
We turn the three sets of traces (training, validation, and test) into three sets of event prefixes by taking the event prefixes for each length $1 < k < |\sigma|$ per trace.
In contrast to~\cite{navarin_lstm_2017,bukhsh_processtransformer_2021}, we excluded event prefixes of length 1 because a minimum of two events is required to form the graph representation of an event prefix. Moreover, similar to \cite{navarin_lstm_2017}, we excluded complete prefixes (i.e., $k = |\sigma|$) because predicting the remaining time for such prefixes often lacks practical value.

\mypar{Training setup and configuration choices}
We used the AdamW optimizer \cite{loshchilov_decoupled_2019} and cosine with warm-up learning rate scheduling \cite{He_2019_CVPR} to train our model. Training spanned 600 epochs, including 50 warm-up epochs, with a base learning rate of 0.001 and weight decay of $e^{-5}$. We used a batch size of 128, with occasional adjustments for specific event logs. Similarly, the number of training epochs is adjusted to account for variations in validation loss behaviours across different event logs. The key configuration choices in our experiments include:
\begin{compactitem}
    \item \textit{PE/SE modules:} LapPE+RWSE \cite{kreuzer_rethinking_2021,dwivedi_graph_2022} serves as the default module, occasionally substituted by Graphormer \cite{ying_transformers_2021}. By default, DeepSet \cite{zaheer_deep_2017} processes PE/SEs, with SignNet \cite{lim_sign_2022} replacing LapPE+RWSE and DeepSet in some experiments. PE/SEs are tested in two sizes: 8 and 16.

    \item \textit{Embedding modules:} Nodes and edges use an embedding size of 64. Edge features are compressed using two fully-connected layers, though in some experiments we opt for a single layer.
    
    \item \textit{Processing modules:} comprising 5 GPS layers (GIN + Transformer) \cite{rampasek_recipe_2022} with 8 heads, utilizing a dropout of 0.0 for MPNN blocks and 0.5 for Transformer blocks. In some experiments, we used 10 GPS layers with 4 heads instead, while in others we applied a dropout of 0.2 for MPNN blocks. 

    \item \textit{Readout layer:} Mean pooling is the default configuration, occasionally replaced by sum pooling.
\end{compactitem}
\noindent Our focus is on demonstrating PGTNet's applicability for remaining time prediction rather than an exhaustive hyperparameter search. Therefore, we evaluated a limited set of configurations per log, selecting the best based on validation loss.

\mypar{Evaluation metrics}
We used Mean Absolute Error (MAE) to measure prediction accuracy. 
Since  we are interested in models that not only have smaller MAE but also can make accurate predictions earlier, allowing more time for corrective actions,  we used the method proposed in \cite{tax_predictive_2017}, which evaluates MAE across different event prefix lengths. This approach provides insights into the predictive performance of the model as more events arrive.

\subsection{Results}
\label{subsec:results}

\mypar{Overall results} 
\autoref{tab2} summarizes the experimental results  for the 20 event logs, providing the average and standard deviations of the MAEs obtained over experiments with three distinct, random seeds for training and evaluation. The table shows that our approach consistently outperforms the benchmark approaches across the 20 event logs, yielding an average MAE of 12.92, compared to 24.63 for the next best approach (GGNN).

\begin{table}[ht]
\caption{Mean Absolute Error for remaining time prediction (MAE: in days).}\label{tab2}
\robustify\bfseries
\centering
\sisetup{text-series-to-math}
\begin{tabular}{
    @{}
    l
    S[table-format=3.2,
    table-number-alignment = right]
    S[table-format=3.2,
    table-number-alignment = right,
    separate-uncertainty = true,
    table-figures-uncertainty = 2]
    S[table-format=3.2,
    table-number-alignment = right,
    separate-uncertainty = true,
    table-figures-uncertainty = 2]
    S[table-format=3.2,
    table-number-alignment = right,
    separate-uncertainty = true,
    table-figures-uncertainty = 2]
    S[table-format=3.2,
    table-number-alignment = right,
    separate-uncertainty = true,
    table-figures-uncertainty = 2]
    @{}
    }
\hline
Event log &
\multicolumn{1}{m{1.1cm}}{\centering DUMMY} &
\multicolumn{1}{m{2.1cm}}{\raggedleft DALSTM} &
\multicolumn{1}{m{2.0cm}}{\centering Process Transformer} &
\multicolumn{1}{m{1.8cm}}{\centering GGNN} &
\multicolumn{1}{m{2.1cm}}{\centering PGTNet} \\
\hline
    Env.permit & 5.21 & 3.36\pm0.04 & 4.26\pm0.04 & 3.52\pm0.02 & \bfseries2.72\pm0.08 \\
    Helpdesk & 9.15 & 8.22\pm0.23 & 6.33\pm0.01 & 6.21\pm0.04 & \bfseries 4.11\pm0.04 \\    
    BPIC12 & 9.03 & 9.34\pm0.41 & 7.11\pm0.02 & 4.78\pm0.01 & \bfseries2.31\pm0.19 \\
    BPIC12W & 9.16 & 8.22\pm0.06 & 7.40\pm0.01  & 5.12\pm0.02 & \bfseries2.70\pm0.01 \\
    BPIC12C & 8.92 & 8.21\pm0.27 & 6.86\pm0.01 & 5.32\pm0.01 & \bfseries2.77\pm0.02 \\  
    BPIC12CW & 9.17 & 8.04\pm0.09 & 7.46\pm0.01 & 6.99\pm0.01 & \bfseries 5.07\pm0.03 \\    
    BPIC12O & 8.39 & 8.21\pm0.09 & 7.29\pm0.01 & 6.93\pm0.04 & \bfseries5.57\pm0.01 \\    
    BPIC12A & 8.17 & 7.62\pm0.03 & 7.79\pm0.01 & 7.48\pm0.01 & \bfseries7.38\pm0.01 \\   
    BPIC20I & 27.20 & 20.43\pm0.39 & 17.06\pm0.11 & 15.67\pm0.04 & \bfseries7.67\pm0.19 \\    
    BPIC20D & 4.33 & 4.15\pm0.12 & 3.65\pm0.01 & 3.25\pm0.01 & \bfseries 3.10\pm0.01 \\
    Sepsis & 41.12 & 25.21\pm0.66 & 34.77\pm0.18 & 19.44\pm0.05 & \bfseries16.48\pm0.19 \\
    Hospital & 59.41 & 43.66\pm0.10 & 47.00\pm0.07 & 41.84\pm0.06 & \bfseries 35.68\pm0.03 \\ 
    BPIC15-1 & 50.22 & 36.48\pm2.69 & 31.01\pm0.36 & 16.77\pm0.01 & \bfseries1.76\pm0.06 \\   
    BPIC15-2 & 83.11 & 63.66\pm2.36 & 44.04\pm0.48 & 20.76\pm0.05 & \bfseries3.02\pm0.07 \\   
    BPIC15-3 & 28.76 & 17.69\pm1.16 & 15.23\pm0.23 & 7.06\pm0.03 & \bfseries1.54\pm0.23 \\    
    BPIC15-4 & 56.75 & 53.33\pm2.63 & 34.40\pm0.42 & 17.97\pm0.03 & \bfseries1.65\pm0.06 \\
    BPIC15-5 & 45.97 & 42.89\pm3.08 & 27.76\pm0.28 & 13.61\pm0.08 & \bfseries1.61\pm0.01 \\
    BPIC13I & 16.18 & 7.60\pm0.45 & 13.54\pm0.04 & 11.99\pm0.04 & \bfseries2.23\pm0.05 \\
    BPIC13C & 152.93 & 91.82\pm1.48 & 127.01\pm0.85 & 123.28\pm0.53 & \bfseries37.44\pm1.49 \\
    Traffic fines & 196.26 & 187.41\pm0.53 & 187.08\pm0.11 & 154.56\pm0.19 & \bfseries 113.53\pm0.12 \\
    \hline
    Average & 41.47 & 32.78\pm0.84 & 31.85\pm0.16 & 24.63\pm0.06 & \bfseries12.92\pm0.14 \\
    \hline
\end{tabular}
\end{table}

\begin{figure}
  \centering
  \includegraphics[width=1\textwidth]{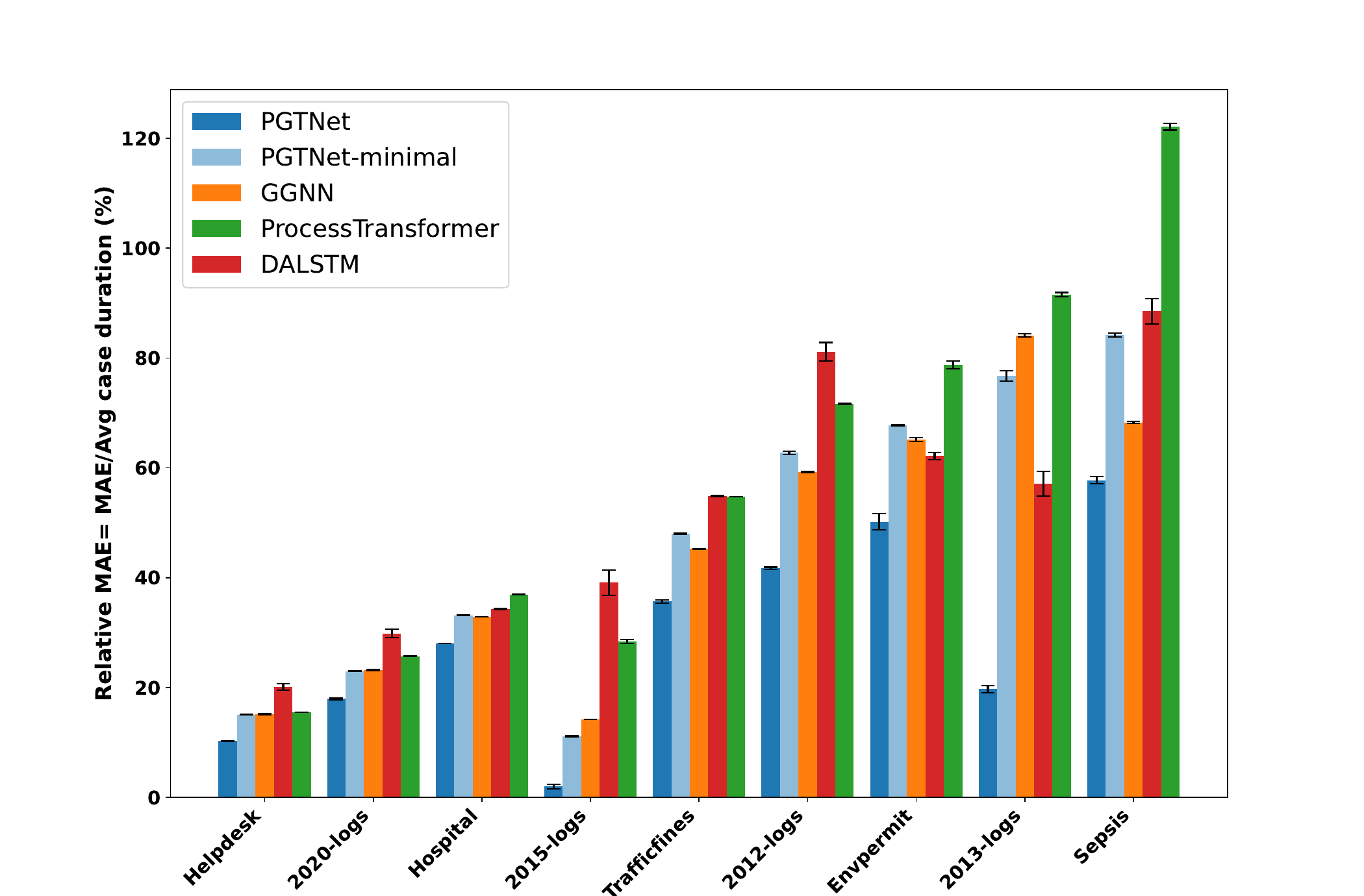}
  \caption{Remaining time prediction accuracy in terms of relative MAE (in percentage).}
  \label{fig:comparison_plot}
\end{figure}

Next to these absolute MAE scores, we also computed the relative MAE (i.e., the MAE divided by the average case duration per log) to account for differences in the cycle times across event logs. 
Using these relative scores, we can visualize the accuracy improvements across different logs, as done in \autoref{fig:comparison_plot} (for clarity, we aggregated the results of logs that stem from the same BPI collection, e.g., averaging the results of all BPIC15 logs).

\begin{figure}[ht]
    \centering    \includegraphics[width=1\linewidth]{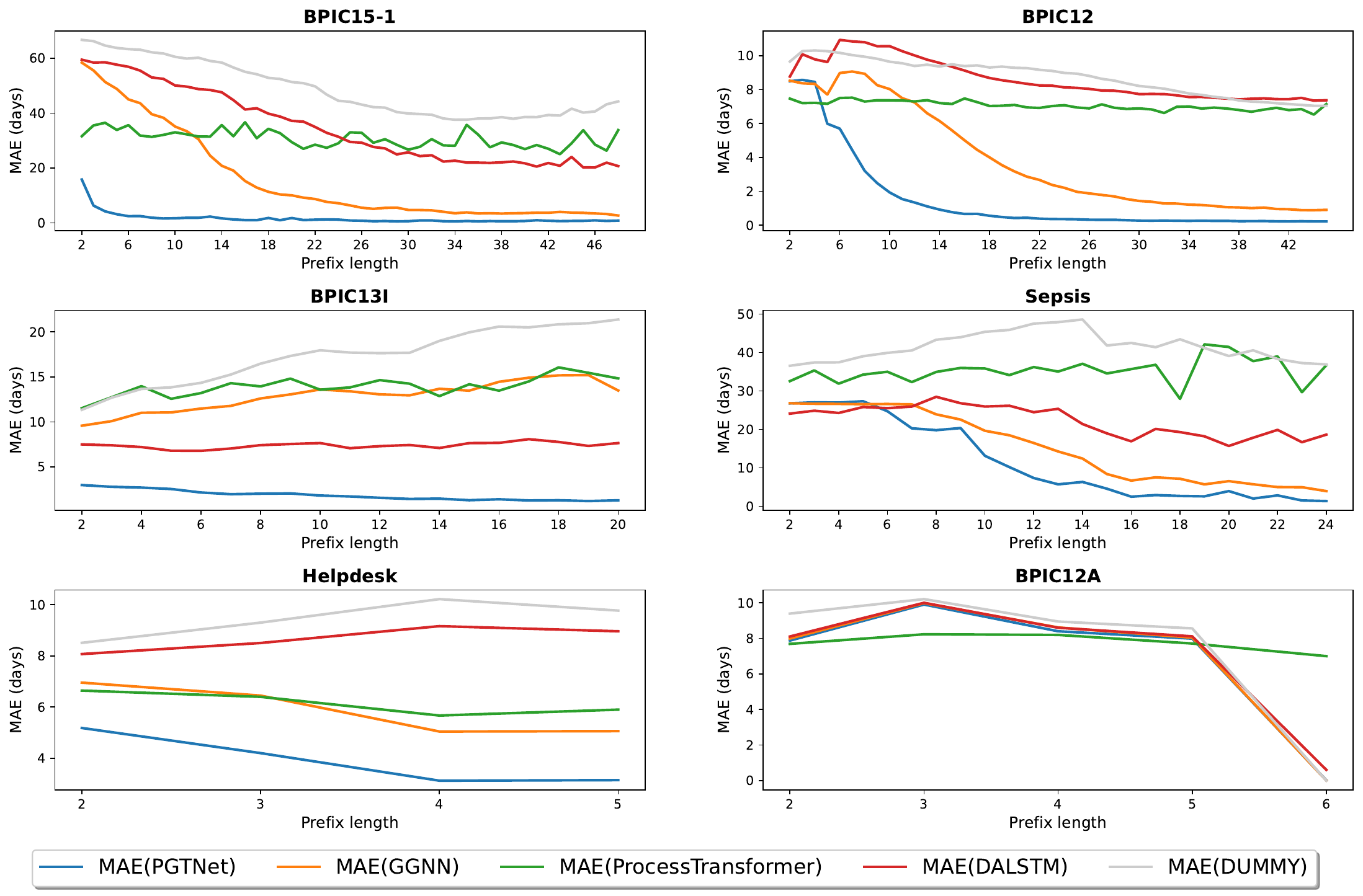}
    \caption{MAE over different prefix lengths (selected event logs).}
    \label{fig:Earliness}
\end{figure}

PGTNet also excels in earliness of predictions, achieving lower MAE for most prefix lengths across all event logs. We illustrate some of these results in \autoref{fig:Earliness}, which depicts MAE trends for \changed{BPIC15-1, BPIC12, BPIC13I, Sepsis, Helpdesk, and BPIC12A logs at various prefix lengths. These six logs are representative of major MAE trends} that we observed across the 20 event logs (the remaining figures are available in our repository). \changed{To improve understandability, we exclude uncommonly long prefixes from the visualization, so that \autoref{fig:Earliness} focuses on prefixes up to a length that corresponds to 90\% of all prefixes in each dataset.}

Summarizing the overall results in terms of MAE, relative MAE, and prediction earliness, we obtain the following main insights:
\begin{compactitem}
  \item \changed{In 11 out of 20 logs, PGTNet achieves an MAE improvement of over 50\% compared to the best baseline approach. For five BPIC15 and two BPIC13 logs, MAE trends closely mirror those of BPIC15-1 and BPIC13I as depicted in \autoref{fig:Earliness}. In BPIC12, BPIC12W, BPIC12C, and BPIC20I,  the MAE trends closely follow those observed in BPIC12. In BPIC15 and BPIC12 logs, ProcessTransformer is the best baseline approach for short prefixes, whereas GGNN exhibits competitive performance only after execution of a substantial number of events. Notably, in the BPICS13 event logs, DALSTM outperforms other baseline approaches.}
  \item In another 7 logs, PGTNet exhibits considerable improvement in MAE (15\% to 35\%), \changed{with Traffic fines, Env.permit and BPIC12O showing similar MAE trends to Helpdesk log in \autoref{fig:Earliness}. MAE trends for BPIC12CW resemble those of other BPIC12 logs in the first group. Notably, the Sepsis log shows a distinct MAE trend, where GGNN achieves comparable results to PGTNet for most prefix lengths.}  
  \item For BPIC20D and BPIC12A, the improvement is more modest, with the latter case yielding nearly identical MAEs for all prefix lengths (see \autoref{fig:Earliness}).   
\end{compactitem}

\mypar{Ablation study}
The remarkable performance of PGTNet can be attributed to a synergy between the expressive capacity of the employed architecture and the incorporation of diverse process perspectives into the graph representation of event prefixes. To distinguish the impacts of these factors, we conducted an ablation study for which we trained a minimal PGTNet model,  relying solely on edge weights (i.e., control-flow) and temporal features, thus omitting data attributes from consideration. We used identical hyperparameters and configurations as was done for the complete PGTNet model. Hence, the ablation study establishes a lower boundary for contribution of the PGTNet's architecture.

Our experiments reveal that the minimal PGTNet model consistently outperforms ProcessTransformer in terms of MAE (see \autoref{fig:comparison_plot}). This underscores PGTNet's capabilities in capturing both local and global contexts, \changed{which is advantageous for predicting remaining time. However, the predictive performance of the minimal PGTNet is comparable to that of the GGNN approach, suggesting that learning from local control-flow structures in MPNN blocks (done by GGNN) holds greater significance than capturing long-range dependencies (as done by ProcessTransformer). This observation is further supported by the overall results presented in \autoref{tab2}, where GGNN outperforms DALSTM and ProcessTransformer in all event logs, except for two BPIC13 logs.} Additionally, the contribution of the architecture and the incorporation of extra data attributes varies across different event logs. While the PGTNet's architecture plays a decisive role for logs such as BPIC15 and BPIC20I, the improvements in MAE for logs such as BPICS12 and BPICS13 is primarily due to the incorporation of additional features. Further details regarding our ablation study can be found in our repository.

\mypar{Impact of process complexity}
The MAE improvement achieved by PGTNet varies significantly across different event logs. Investigating these variations, we correlated process complexity metrics from \cite{augusto_connection_2022, vidgof_impact_2023} with MAE improvements achieved by PGTNet. Notably, our approach outperforms alternatives when the number of process variants increases rapidly with respect to the number of cases. This trend extends to other \textit{variation metrics}, including \textit{`structure'} (average distinct activities per case) and \textit{`level of detail'} (number of acyclic paths in the transition matrix) \cite{augusto_connection_2022}. In terms of \textit{size metrics}, PGTNet exhibits superior performance with increasing average trace length and/or \changed{number of distinct activities}.  The most significant positive correlation is observed for \textit{`normalized sequence entropy'}, a graph entropy metric adept at capturing both variation and size complexity \cite{augusto_connection_2022}.

This reveals that PGTNet excels in highly flexible and complex processes, where \changed{benchmark approaches} may overlook sparse but meaningful directly-follows relations among activities. The graph representation, detailed in \autoref{subsec:transformation}, converts different process variants into non-isomorphic graphs with varying nodes, connectivity structures, and edge weights. Graph isomorphism network (GIN) modules, renowned for distinguishing non-isomorphic graphs \cite{He_2019_CVPR}, process this graph-oriented data. Simultaneously, Transformer blocks capture long-range dependencies. In complex processes like BPIC15, GIN blocks benefit from diverse set of non-isomorphic graphs available for learning, while Transformer blocks leverage insightful PE/SEs, thus providing a synergy that results in a remarkable reduction in MAE. 

\mypar{Training, and inference time} \changed{We conducted experiments using an Nvidia RTX A6000 GPU, with training and inference times detailed in \autoref{tab3}. For training time, we computed the sum of training time for all cross-validation data splits and then averaged these times across all 20 event logs. Regarding average inference time, we compute the time to infer the remaining time per event prefix and report the average inference time across all event logs.}

\changed{In terms of training time, DALSTM and ProcessTransformer, which use shallow neural networks (either with 2 LSTM layers or 1 self-attention layer), can be trained an order of magnitude faster than the graph-based approaches, though PGTNet is still trained faster than GGNN (12.93 vs. 18.56 hours). We see a similar trend in terms of inference times, though essentially all approaches are reasonably fast here, with PGTNet being the slowest with an average time just below 3 miliseconds.}

\begin{table}[ht]
\caption{Average training and inference time for remaining time prediction.}\label{tab3}
\centering
\sisetup{text-series-to-math}
\begin{tabular}{
    @{}
    l
    S[table-format=2.2,
    table-number-alignment = center]
        S[table-format=2.2,
    table-number-alignment = center]
        S[table-format=2.2,
    table-number-alignment = center]
        S[table-format=2.2,
    table-number-alignment = center]
    @{}
    }
\hline
\multicolumn{1}{m{3.0cm}}{Time} &
\multicolumn{1}{m{1.8cm}}{\centering DALSTM} &
\multicolumn{1}{m{1.8cm}}{\centering Process Transformer} &
\multicolumn{1}{m{1.8cm}}{\centering GGNN} &
\multicolumn{1}{m{1.8cm}}{\centering PGTNet} \\
\hline
    Training time (hours) & 1.68 & 1.64 & 18.56 & 12.93 \\
    Inference time (miliseconds) & 0.51 & 0.14 & 2.14 & 2.96 \\ 
\hline
\end{tabular}
\end{table}

\section{Conclusion and Future Work}
\label{sec:conclusion}
This paper introduces a novel approach employing Process Graph Transformer Networks (PGTNet) to predict the remaining time of running business process instances. Our approach consists of a data transformation from an event log to a graph dataset, and training a neural network based on the GPS Graph Transformer recipe \cite{rampasek_recipe_2022}. Our graph representation of event prefixes incorporates multiple process perspectives and also enables integration of control-flow relationships among activities into the learning process. This graph-oriented data input is subsequently processed by PGTNet, which strikes a balance between learning from local contexts and long-range dependencies.

Through experiments conducted on 20 real-world datasets, our results demonstrate the superior accuracy and earliness of predictions achieved by PGTNet compared to the existing deep learning approaches. Notably, our approach exhibits exceptional performance for highly flexible and complex processes, where the performance of LSTM, Transformer \changed{and GGNN} architectures falls short.

While originally designed for predicting remaining times, our approach has the potential to learn high-level event prefix representations, rendering it applicable to other tasks, including next activity prediction and process outcome prediction. 
In future research, we therefore aim to apply PGTNet for these tasks, whereas we also aim to improve the predictive accuracy of our approach by investigating the potential of multi-task learning  and exploring different positional and structural embeddings, as well as varying graph representations.
Finally, we aim to extend our approach to also be applicable to object-centric event logs.

\myparthree{Reproducibility} Our source code and all evaluation results  are accessible in our repository: 
\url{https://github.com/keyvan-amiri/PGTNet}
%
%
\bibliographystyle{splncs04}
\bibliography{references}
\end{document}